\documentclass{article}
\usepackage{ijcai22}

\usepackage{times}
\usepackage{soul}
\usepackage{url}
\usepackage[hidelinks]{hyperref}
\usepackage[utf8]{inputenc}
\usepackage[small]{caption}
\usepackage{graphicx}
\usepackage{amsmath}
\usepackage{amsthm}
\usepackage{amsfonts}
\usepackage{booktabs}
\usepackage{algorithm}
\usepackage{algorithmic}
\title{Multilevel Hierarchical Network with Multiscale Sampling \\for Video Question Answering}

\author{
Min Peng$^{1,2}$\thanks{Equal contribution.}\and
Chongyang Wang$^{3*}$\and
Yuan Gao$^4$\and
Yu Shi$^2$\And 
Xiang-Dong Zhou$^2$\\
\affiliations
$^1$University of Chinese Academy of Sciences\\
$^2$Chongqing Institute of Green and Intelligent Technology, Chinese Academy of Sciences\\
$^3$University College London\\
$^4$Shenzhen Institute of Artificial Intelligence and Robotics for Society\\
\emails
\{pengmin, shiyu, zhouxiangdong\}@cigit.ac.cn,
\{mvrjustid, gaoyuankidult\}@gmail.com
}

\begin{document}

\maketitle

\begin{abstract}
    Video question answering (VideoQA) is challenging given its multimodal combination of visual understanding and natural language processing. While most existing approaches ignore the visual appearance-motion information at different temporal scales, it is unknown how to incorporate the multilevel processing capacity of a deep learning model with such multiscale information. Targeting these issues, this paper proposes a novel Multilevel Hierarchical Network (MHN) with multiscale sampling for VideoQA. MHN comprises two modules, namely Recurrent Multimodal Interaction (RMI) and Parallel Visual Reasoning (PVR). With a multiscale sampling, RMI iterates the interaction of appearance-motion information at each scale and the question embeddings to build the multilevel question-guided visual representations. Thereon, with a shared transformer encoder, PVR infers the visual cues at each level in parallel to fit with answering different question types that may rely on the visual information at relevant levels. Through extensive experiments on three VideoQA datasets, we demonstrate improved performances than previous state-of-the-arts and justify the effectiveness of each part of our method.
\end{abstract}

\section{Introduction}
    With the advancements of deep learning in computer vision and natural language processing~\cite{1,2,3}, Video Question Answering (VideoQA) has lately received more attention for its wide application in video retrieval, intelligent QA system, and autonomous driving. In comparison with Image Question Answering (ImageQA) ~\cite{4,5}, VideoQA is more difficult because it needs to properly extract the dynamic interaction between the text and the video in addition to modeling the semantic association between the text and a single image.

\begin{figure}[t]
\centering
\includegraphics[width=0.47\textwidth]{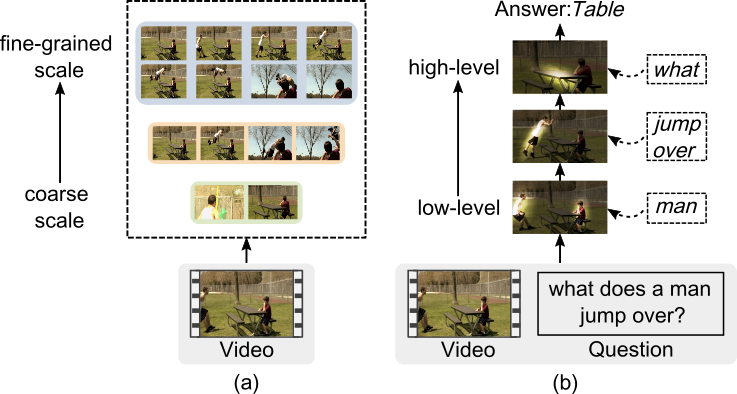}
\caption{(a) The multiscale property of a video example, where at a fine-grained scale the richer frames contribute to understanding general action and logical information, and the local attributes could be better inferred with fewer frames at a coarser scale. (b) The typical multilevel processing of a deep learning model, where the increase of feature levels leads to the transition of learning from local objects to global semantics.}
\label{fig1}
\end{figure}

    The majority of existing methods ~\cite{13,17,18,19,20,25,26} used recurrent neural networks (RNNs) and their variants to connect the embeddings of the text and spatial features extracted with convolutional neural networks (CNNs) of the video, and adopted spatial-temporal attention mechanisms to learn the text-related visual representation ~\cite{9,10} or the so called co-attention representation ~\cite{11,12}. To acquire the long-term interaction between the question and the video, some methods \cite{15,16} proposed to use extra memory modules to augment the capacity of sequential encoding. While these methods achieved interesting results on benchmark datasets of VideoQA, the multiscale semantic relations existed between the text and appearance-motion information of the video is largely ignored. 
    
    For the video example shown in Fig.\ref{fig1} (a), the local attribute of the `man' is better inferred at a coarser scale, while the semantic information of `what' and `jump over' is mostly revealed from consecutive frames at a finer-grained scale. For such a multiscale characteristic, a model should be able to search sufficient information from the video given different question types that rely on the visual clues at different scales. In addition, as illustrated in Fig.\ref{fig1} (b), the multilevel representation learning of a deep learning model covers generalizable information of local objects and global semantics of the input video along the increase of model depth \cite{27}. It remains an open question about how to incorporate the multiscale information of a video with multilevel processing of a deep learning model for VideoQA.

\begin{figure*}[t]
\centering
\includegraphics[width=0.98\textwidth]{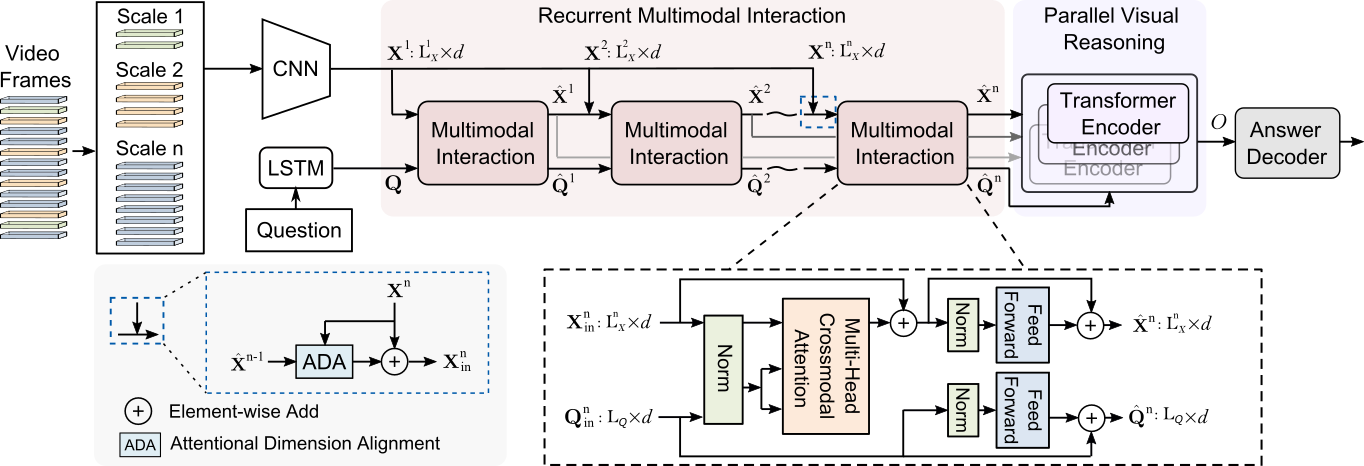}
\caption{An overview of our MHN method. With a multiscale sampling, a series of frame groups $\mathbf{C}^n$ are produced, from which the multiscale appearance-motion features $\mathbf{X}^n$ are extracted with CNNs. The RMI module uses a recurrent structure to incorporate its multilevel processing capacity with such multiscale input. Thereon, the PVR module establishes the visual reasoning per each level in parallel, and fuses the output visual semantics $\hat{\mathbf{X}}^n$ under the guidance of a high-level textual representation $\hat{\mathbf{Q}}^n$ to produce the answering feature $O$.}
\label{fig2}
\end{figure*}

    Given the above findings, we propose a novel method named Multilevel Hierarchical Network (MHN) with multiscale sampling for VideoQA, as shown in Fig.\ref{fig2}. MHN comprises two modules of Recurrent Multimodal Interaction (RMI) and Parallel Visual Reasoning (PVR). To leverage the multiscale visual information, we first apply a multiscale sampling to acquire several frame groups from the input video. To accommodate the frame groups at different scales, the RMI module uses a recurrent structure to bridge the multimodal interaction blocks, where each block takes the frame group at a scale as an input. In the later section, we empirically analyze the impact of the ways of matching the scales of frame groups with the levels of different multimodal interaction blocks within this module. Each multimodal interaction block extracts the question-guided visual representation per scale, and the recurrent structure provides the representations across different levels. The PVR module takes this output at each level for visual reasoning, where a transformer encoder is shared during the parallel processing. In this way, our method fits with different question types, where their answering could benefit from the visual clues at relevant levels.

    Our contributions are as follows: 1) We propose a novel Multilevel Hierarchical Network (MHN) with multiscale sampling for VideoQA, to incorporate the multiscale interaction between the text and the video with the multilevel processing capability of a deep learning model; 2) We design a recurrent multimodal interaction module to enable the multimodal multilevel interaction between the two input modalities, and a parallel visual reasoning module to infer the visual clues per each level; 3) We conduct comprehensive evaluations on TGIF-QA, MSRVTT-QA, and MSVD-QA datasets, achieving improved performances than previous state-of-the-arts and verifying the validity of each part of our method.

\section{Related Work}
    
    \paragraph{VideoQA} challenges a model on analyzing the complex interaction between the text and visual appearance-motion information. ~\cite{9} proposed a method based on Gradually Refined Attention to extract the appearance-motion features using the question as guidance. ~\cite{10} proposed a dual-LSTM approach together with spatio-temporal attention to extract visual features. Later on, other spatio-temporal attention-based methods proposed to use co-attention representation ~\cite{11,12}, hierarchical attention network ~\cite{14}, and memory-augmented co-attention models ~\cite{15,16} for the extraction of motion-appearance features and question-related interactions. Recently, ~\cite{17} proposed a multistep progressive attention model to prune out irrelevant temporal segments, and a memory network to progressively update the cues to answer. Additionally, some proposed to leverage object detection across the video frames to acquire fine-grained appearance-question interactions ~\cite{13,18}. ~\cite{19} proposed to use a hierarchical structure for the extraction of question-video interactions from the frame-level and segment-level. ~\cite{20} proposed a heterogeneous multimodal graph structure using the question graph as an intermediate bridge to extract the internal semantic relation between each word and the video. Although interesting results are achieved by these methods, the multiscale structure of the answering cues existed in the visual information is not well explored.
    
    \paragraph{Transformer} ~\cite{2} has achieved outstanding performance using its self-attention mechanism and forward-passing architecture, which is first introduced for neural machine translation tasks. Given its promising efficiency in analyzing temporal information, transformer becomes one of the dominant approaches for various NLP tasks ~\cite{21}. Recent efforts are seen in transferring transformer to the computer vision domain, where improved performances are achieved in object detection ~\cite{22}, instance segmentation ~\cite{23}, and action recognition ~\cite{24}. In this paper, we propose to use a shared transformer encoder to extract the semantic interaction between the question and the video at different levels in parallel for visual inference.

\section{Method}
    Given a video $\mathcal{V}$ and the question $\mathcal{Q}$, VideoQA aims to acquire the correct answer $\hat{a}$. For open-ended and multi-choice types of question, the answer space $\mathcal{A}$ comprises the group of pre-defined answers and list of candidate answering options, respectively. Generally, VideoQA is formulated as follows.
\begin{equation}
    \hat{a}=\mathop{\arg\max}\limits_{a\in\mathcal{A}}f_\theta(a\mid \mathcal{Q},\mathcal{V}),
\label{equa1}
\end{equation}
    where $\theta$ represents the group of trainable parameters of the modeling function $f$. 
    
    As shown in the overview of our proposed MHN model in Fig.\ref{fig2}, we first extract multiscale appearance-motion features from the input video, and embeddings from the input question. Thereon, the proposed Recurrent Multimodal Interaction (RMI) module propagates the information across its multimodal interaction blocks in a recurrent manner, providing the semantic representations at different levels. Note, in this section, we assume that these blocks from the low-level to the high-level accommodate the visual information from coarser scale (fewer frames) to finer-grained scale (richer frames) accordingly. Thus, the scale and level $n$ are the same in this case. With such multilevel representations, the proposed Parallel Visual Reasoning (PVR) module uses a shared transformer block to establish the visual inference at each level and produces the final visual cues after a fusion operation. Finally, a classification or regression operation is done in the decoder for answering. It is worth mentioning that, similar to the methods we compare with in this paper, our method does not rely on large-scale pre-training and big models to achieve the improved performances.
    
\subsection{Multiscale Sampling and Feature Extraction}

    \noindent\textbf{Multiscale Sampling}. Different from some previous VideoQA works ~\cite{13,16,19,20} that adopted dense sampling for the input video, we conduct a multiscale sampling to help acquire visual features at different temporal scales. For input video $\mathcal{V}$, at scale $n\in\{1,...,N\}$, we sample $T\times2^{n-1}$ frames along the forward temporal direction, with $T$ as the size of our sampling window, which is set to $16$ in our experiment. At this scale, for a clip comprising $T$ frames, the group of such clips $\mathbf{C}^n$ is represented as
\begin{equation}
    \mathbf{C}^n=\lbrace \mathbf{c}^{n,1}, \mathbf{c}^{n,2},...,\mathbf{c}^{n,2^{n-1}} \rbrace,
\label{equa2}
\end{equation}
    where $\mathbf{c}^{n,i}$ is the $i$-th video clip sampled at scale $n$. Given the same number of video clips sampled from the video, multiscale sampling provides richer visual information than dense sampling, covering more aspects from the local object to the global interaction or event, which help the network to better exert its multilevel processing capacity on understanding the semantic relations between the question and the video.

    \noindent\textbf{Visual Representation}. Following ~\cite{11,13,15,16,18,19,20}, we use ResNet ~\cite{1} and 3D ResNet152 ~\cite{3} to extract the appearance and motion features, respectively, from the video. Therein, the feature output at the last pooling layer of each network is used. For the group of video clips $\mathbf{C}^n$ at scale $n$, the extracted frame-wise appearance feature $\mathbf{V}^n$ can be represented as
\begin{equation}
    \mathbf{V}^n=\lbrace v^{n,i} \mid v^{n,i} \in \mathbb{R}^{2048} \rbrace^{T\times2^{n-1}}_{i=1}.
\label{equa3}
\end{equation}

    Similarly, we extract the clip-wise motion feature $\mathbf{M}^n$ as
\begin{equation}
    \mathbf{M}^n=\lbrace m^{n,t} \mid m^{n,t} \in \mathbb{R}^{2048} \rbrace ^{2^{n-1}}_{t=1}.
\label{equa4}
\end{equation}

    We use a linear feature transformation layer to map feature vectors in $\mathbf{V}^n$ and $\mathbf{M}^n$ into a $d$-dimensional feature space, where we now have $v^{n,i},m^{n,t}\in\mathbb{R}^{d}$. After a feature concatenation following the temporal order of video clips, the appearance-motion feature $\mathbf{X}^n$ at $n$-th scale is represented as
\begin{equation}
    \mathbf{X}^n=\lbrace x^{n}_j \mid x^{n}_j \in \mathbb{R}^{d} \rbrace ^{\mathrm{L}^n_X}_{j=1},
\label{equa5}
\end{equation}
    where $x^{n}_j\in\lbrace v^{n,i},m^{n,t} \rbrace$, with the total number of appearance-motion features at scale $n$ being $\mathrm{L}^n_X=2^{n-1}(T+1)$. Meanwhile, such feature is added with the positional embedding $\mathbf{P}^n\in\mathbb{R}^{\mathrm{L}_X^n \times d}$, in order to maintain the positional information of the feature sequence.
    
    \noindent\textbf{Linguistic Representation}. For the multi-choice question and answer candidates, we adopt Glove word embedding method ~\cite{6} to acquire the $300$-dimensional feature embeddings, which is further mapped into a $d$-dimensional space using linear transformation layers. Thereon, a bidirectional LSTM network is adopted to extract the contextual semantics between each word in the question, and the answer, respectively. Finally, the acquired representation for the question or answer candidates is obtained by concatenating the hidden states of the last forward and backward LSTM layer per timestep, written as
\begin{equation}
    \mathbf{Q}=\lbrace q_j \mid q_j \in \mathbb{R}^d \rbrace^{\mathrm{L}_Q}_{j=1},
\label{equa6}
\end{equation}
and
\begin{equation}
    \mathbf{A}^k=\lbrace a^k_j \mid a^k_j \in \mathbb{R}^d \rbrace^{\mathrm{L}_A^k}_{j=1},
\label{equa7}
\end{equation}
    where $\mathrm{L}_Q$ and $\mathrm{L}_A^k$ are the number of words in the question and $k$-th answer candidate, respectively.
    
\subsection{Recurrent Multimodal Interaction}
    Within the RMI module, the recurrent connections of multimodal interaction blocks facilitate the extraction of multilevel semantic relations between the question and the video. In a block that processes information at scale $n$, with the input visual feature $\mathbf{X}^n_{\mathrm{in}}$ and question feature $\mathbf{Q}^n_{\mathrm{in}}$, the interim interaction output $\tilde{\mathbf{X}}^n$ is computed as
\begin{equation}
    \tilde{\mathbf{X}}^n=\mathbf{X}^n_{\mathrm{in}}+\mathrm{MCA}(\mathbf{X}^n_{\mathrm{in}},\mathbf{Q}^n_{\mathrm{in}}),
\label{equa8}
\end{equation}
    where $\mathrm{MCA}(\cdot)$ denotes the operation in a multi-head crossmodal attention layer, and at a single attention head $h$ its output is
\begin{equation}
    \mathrm{MCA}^h=\mathrm{softmax}(\frac{\mathbf{F}^h_q\mathbf{F}_k^{h\top}}{\sqrt{d}})\mathbf{F}^h_v,
\label{equa9}
\end{equation}
    where $\mathbf{F}_q^h=\mathrm{LN}(\mathbf{X}^n_{\mathrm{in}})\mathbf{W}_q^h$ is the Query, $\mathbf{F}_k^h=\mathrm{LN}(\mathbf{Q}^n_{\mathrm{in}})\mathbf{W}_k^h$ is the Key, and $\mathbf{F}_v^h=\mathrm{LN}(\mathbf{Q}^n_{\mathrm{in}})\mathbf{W}_v^h$ is the Value, with $\mathrm{LN}(\cdot)$ being the layer normalization ~\cite{7}. We concatenate the output per head to obtain
\begin{equation}
    \small \mathrm{MCA}(\mathbf{X}^n_{\mathrm{in}},\mathbf{Q}^n_{\mathrm{in}})=\mathrm{concat}(\mathrm{MCA}^1,\mathrm{MCA}^2,...,\mathrm{MCA}^H),
\label{equa10}
\end{equation}
    where $\mathbf{W}_q^h$, $\mathbf{W}_k^h$, $\mathbf{W}_v^h\in\mathbb{R}^{d\times d/H}$, and $\mathbf{W}_o\in\mathbb{R}^{d\times d}$ are the learnable weight matrices, and $H$ is the total number of attention heads. In short, within the $\mathrm{MCA}$, the question semantics and appearance-motion features are connected, while the semantic co-occurrence of them is extracted using the attention mechanism.
    
    Given $\tilde{\mathbf{X}}^n$ and $\mathbf{Q}^n_{\mathrm{in}}$, a feed forward layer is further added to each modality to acquire the final interaction feature outputs $\hat{\mathbf{X}}^n\in\mathbb{R}^{\mathrm{L}_X^n \times d}$ and $\hat{\mathbf{Q}}^n\in\mathbb{R}^{\mathrm{L}_Q \times d}$ as
\begin{equation}
    \hat{\mathbf{X}}^n=\tilde{\mathbf{X}}^n+f_X(\mathrm{LN}(\tilde{\mathbf{X}}^n)),
\label{equa11}
\end{equation}
    and
\begin{equation}
    \hat{\mathbf{Q}}^n=\mathbf{Q}^n_{\mathrm{in}}+f_Q(\mathrm{LN}(\mathbf{Q}^n_{\mathrm{in}})),
\label{equa12}
\end{equation}
    where $f_X(\cdot)$ and $f_Q(\cdot)$ represent the operation in a feed forward layer, which comprises two linear projections separated by a GELU non-linearity. The feature dimension $d$ stays unchanged.
    
    We design a \textit{recurrent connection} to connect multimodal interaction blocks at different levels for multilevel processing, which also handles the difference in temporal dimension of the feature at each level. Given the interaction feature output $\hat{\mathbf{X}}^{n-1}$ of the previous block and current appearance-motion feature input $\mathbf{X}^n$, the recurrent connection (shown in the area marked by blue dashed contour in Fig.\ref{fig2}) uses an attentional dimension alignment to provide the input $\mathbf{X}^n_{\mathrm{in}}$ for the current block as
\begin{equation}
    \small \mathbf{X}^n_{\mathrm{in}}=\mathbf{X}^n+\mathrm{softmax}((\mathbf{X}^n\mathbf{W}_1^{n-1})(\hat{\mathbf{X}}^{n-1}\mathbf{W}_2^{n-1}))\hat{\mathbf{X}}^{n-1},
\label{equa13}
\end{equation}
    where $\mathbf{W}_1^{n-1}$ and $\mathbf{W}_2^{n-1}$ are learnable weight matrices. In addition, we have $\mathbf{Q}_{\mathrm{in}}^{n}=\hat{\mathbf{Q}}^{n-1}$. The extra bypass scheme within the connection ensures that the current block receives the higher-level information. For the whole module, we repeat Equation \ref{equa13} and Equation \ref{equa8}-\ref{equa12} to propagate the vision-question interaction recurrently across different levels.
    
\subsection{Parallel Visual Reasoning}
    An essential step of VideoQA is to infer the visual cues from the appearance-motion feature via understanding the question semantics. Given vision-question outputs at different levels from the RMI module, the proposed PVR module first infer visual cues at each level and acquire the final feature for answering under the guidance of question semantics.
    
    Given the interaction feature output $\hat{\mathbf{X}}^n$ at scale $n$ (at the highest level) of RMI module, we use the encoder layer proposed in transformer ~\cite{2} to acquire the visual cue $\mathbf{R}^n$ as
\begin{equation}
    \mathbf{R}^n=\mathbf{Z}^n+f(\mathrm{LN}(\mathbf{Z}^n)),
\label{equa14}
\end{equation}
with
\begin{equation}
    \mathbf{Z}^n=\hat{\mathbf{X}}^n+\mathrm{MCA}(\hat{\mathbf{X}}^n,\hat{\mathbf{X}}^n),
\label{equa15}
\end{equation}
    where $f(\cdot)$ denotes the operation in a feed forward layer. For the outputs from RMI module at different scales, PVR module repeats Equation \ref{equa14} and \ref{equa15} with the shared processing layers. This help to maintain the consistency of the semantic space during visual inference, and a compact learnable weight even if the number of scales increases. 

    Given the high-level question feature $\hat{\mathbf{Q}}^n$ output by RMI module, PVR module further fuses the multilevel visual cues to acquire the final feature for answering as

\begin{equation}
    \alpha^n=\mathrm{softmax}(\bar{\mathbf{Q}}^n\bar{\mathbf{R}}^{n\top}),
\label{equa16}
\end{equation}
with
\begin{equation}
    O=\sum^{N}_{n=1}\alpha^n\bar{\mathbf{R}}^n,
\label{equa17}
\end{equation}
    where $\bar{\mathbf{Q}}^n, \bar{\mathbf{R}}^{n}$ are acquired by applying average pooling along the temporal dimension on $\hat{\mathbf{Q}}^n, \mathbf{R}^n$, respectively, and the final feature for answering is $O\in\mathbb{R}^d$.

\subsection{Answer Decoder and Loss Function}
    Following ~\cite{10,11,13,15,16,18,19,20,25,26}, different decoding strategies are used according to the types of question. 

    Specifically, we treat an open-ended question as a multi-class classification task, where the answer decoder aims to predict the correct category from the answer space $\mathcal{A}$. Given the final feature ${O}$, the probability vector $\mathbf{P}\in\mathbb{R}^{|\mathcal{A}|}$ towards each class is computed as 
\begin{equation}
    y=\delta(\mathbf{W}^o{O}+b^o),
\label{equa18}
\end{equation}
\begin{equation}
    \mathbf{P}=\mathrm{softmax}(\mathbf{W}^yy+b^y),
\label{equa19}
\end{equation}
    where $\mathbf{W}^o$, $\mathbf{W}^y$, $b^o$, and $b^y$ are the learnable weight matrices and biases of each layer, respectively; $\delta(\cdot)$ is the activation function. The cross-entropy loss is used here.
    
    For the repetition count task, linear regression is used to replace the classification function shown in Equation \ref{equa21}, the output of which is processed by a rounding function to acquire the integer output. The loss function used here is the Mean Squared Error (MSE).
    
    For multi-choice questions, we use an answer candidate $\mathcal{A}^k$ as input to the MHN model, similar with the question $\mathcal{Q}$. Therein, the learnable parameters are shared for the processing of the answer and the question. Given the final feature outputs ${O}$ conditioned by the question and ${O}^k_a$ conditioned by the $k$-th answer candidate, the predicted probability towards the $k$-th answer candidate is computed as
\begin{equation}
    y^k=\delta(\mathbf{W^*}[\bar{O};\bar{O}^k_a]+b^*).
\label{equa20}
\end{equation}
\begin{equation}
    p^k=\mathbf{W}^*y^k+b^*.
\label{equa21}
\end{equation}
    
    The answer candidate that produces the highest probability $p$ is selected as the predicted for the question. Hinge loss \cite{29}, namely $\mathrm{max}(0,1+p^i-p^{c})$, is adopted to compute the loss between the correct answer $p^{c}$ and the incorrect answer $p^i$.

\section{Experiment}
    
\subsection{Datasets}
    Three VideoQA benchmarks are adopted for our evaluation.
    
    \noindent\textbf{TGIF-QA} \cite{10} is a large-scale dataset for videoQA, which comprises 165K question-answer pairs and 72K animated GIFs. This dataset has four task types, including \textit{Action}, \textit{Transition} (Trans.), \textit{FrameQA}, and \textit{Count}. Action is a multi-choice task aimed to identify the repetitive actions. Trans. is another multi-choice task for identifying the transition actions before or after a target action. FrameQA is an open-ended task where the answer could be inferred from a single frame of the video (GIF file). Count is to count the number of a repetitive action.

    \noindent\textbf{MSVD-QA} \cite{9} comprises 1,970 short clips and 50,505 question-answer pairs, which are divided into five question categories of \textit{what}, \textit{who}, \textit{how}, \textit{when}, and \textit{where}. All of them are open-ended.

    \noindent\textbf{MSRVTT-QA} \cite{9} comprises 10K videos and 243K question-answer pairs. The question types are similar to what included in the MSVD-QA dataset. However, the scenario of the video is more complex, with a longer duration of 10-30 seconds.
    
\subsection{Implementation Details}
\subsubsection{Metrics}
    For multi-choice and open-ended tasks, we use accuracy to evaluate the performance. For the Count task in TGIF-QA dataset, we evaluate with Mean Squared Error (MSE) between the predicted answer and the ground truth.

\subsubsection{Training Details}
    We use the official split of training, validation, and testing sets of each dataset. By default, the maximum scale $N$ is set to 3, and the visual features are input to the model with increasing scales. For each multimodal interaction block in RMI module, the feature dimension $d$ is set to 512, and the number of attentional heads $H$ is set to 8. The number of mini batch size is set to 32, with a maximum number of epochs set to 20. The Adam \cite{8} optimizer is used, with the initial learning rate set to 1e-4, which reduces by half when the loss stops decreasing after every 10 epochs. We implement the method with PyTorch deep learning library on a PC with two GTX 1080 Ti GPUs.
    
\subsection{Comparison with the State-of-the-arts}
    On the TGIF-QA dataset, we compare with a series of state-of-the-art VideoQA methods. As shown in Table \ref{tab:table1}, our MHN model outperforms other state-of-the-art methods across all four tasks. Our improvements are more obvious on Action, Trans., and Count tasks, where the answering requires visual inference at different temporal scales and processing levels of the model. These show the advantage of the multiscale multilevel processing capacity of our model. In addition, most methods use dense sampling for the input video, \textit{e.g.}, HCRN ~\cite{19} and Bridge2Answer ~\cite{20} sampled 8 clips each comprising 16 frames, while our method with scale $N$ set to 3 only samples 7 clips so that costing less computational loads.
    
    Further comparisons on the MSVD-QA and MSRVTT-QA datasets are conducted. Results are reported in Table \ref{tab:table2}. On such more challenging data, our MHN model still achieves the best performances of 40.4\% and 38.6\% on both datasets, respectively. While Bridge2Answer~\cite{20} additionally extracted semantic dependencies from the question using a NLP tool and HOSTR~\cite{13} applied Fast R-CNN for object detection per frame, our model is able to produce even higher performances without such complex feature pre-processing.
    
\begin{table}[t]
\centering
\resizebox{0.48\textwidth}{!}{%
\begin{tabular}{lrrrr}
\toprule
Method                     & Action        & Trans.         & FrameQA       & Count $\downarrow$\\ 
\midrule
ST-TP~\cite{10}         & 62.9          & 69.4          & 49.5          & 4.32 \\
Co-Mem~\cite{25}        & 68.2          & 74.3          & 51.5          & 4.10 \\
PSAC~\cite{11}          & 70.4          & 76.9          & 55.7          & 4.27 \\
HME~\cite{16}           & 73.9          & 77.8          & 53.8          & 4.02 \\
FAM~\cite{15}           & 75.4          & 79.2          & 56.9          & 3.79 \\
L-GCN~\cite{18}         & 74.3          & 81.1          & 56.3          & 3.95 \\
HGA~\cite{28}           & 75.4          & 81.0          & 55.1          & 4.09 \\
HCRN~\cite{19}          & 75.0          & 81.4          & 55.9          & 3.82 \\
Bridge2Answer~\cite{20} & 75.9          & 82.6          & 57.5          & 3.71 \\ 
HOSTR~\cite{13}         & 75.0          & 83.0          & 58.0          & 3.65 \\
\midrule
\textbf{MHN (ours)} & \textbf{83.5} & \textbf{90.8} & \textbf{58.1} & \textbf{3.58} \\ 
\bottomrule
\end{tabular}
}
\caption{Comparison with state-of-the-art methods on TGIF-QA dataset. For the Count task, the lower is better.}
\label{tab:table1}
\end{table}

\begin{table}[t]
\centering
\resizebox{0.48\textwidth}{!}{%
\begin{tabular}{lrr}
\toprule
Method                     & MSVD-QA & MSRVTT-QA \\ 
\midrule
AMU~\cite{9}             & 32.0                & 32.5 \\
HRA~\cite{26}            & 34.4                & 35.0 \\
Co-Mem~\cite{25}         & 31,7                & 31.9 \\
HME~\cite{16}            & 33.7                & 33.0 \\
FAM~\cite{15}            & 34.5                & 33.2 \\
HGA~\cite{28}            & 34.7                & 35.5 \\
HCRN~\cite{19}           & 36.1                & 35.6 \\
Bridge2Answer~\cite{20}  & 37.2                & 36.9 \\
HOSTR~\cite{13}          & 39.4                & 35.9 \\
\midrule
\textbf{MHN (ours)}     & \textbf{40.4}       & \textbf{38.6} \\ 
\bottomrule
\end{tabular}
}
\caption{Comparison with state-of-the-art methods on MSVD-QA and MSRVTT-QA datasets.}
\label{tab:table2}
\end{table}

\subsection{Ablation Study}
    Here, we run several ablation experiments on the TGIF-QA dataset for in-depth analysis of our method. We adopt the default MHN model used above as the baseline.

\begin{table}[t]
\centering
\resizebox{0.48\textwidth}{!}{%
\begin{tabular}{lrrrr}
\toprule
Model              & Action        & Trans.        & FrameQA       & Count $\downarrow$ \\ 
\midrule
MHN w/ single scale, $n$=1       & 76.2          & 80.6          & 57.9          & 4.24               \\
MHN w/ single scale, $n$=2       & 76.7          & 80.6          & 57.7          & 3.99               \\
MHN w/ single scale, $n$=3       & 77.0          & 81.0          & 57.7          & 3.66               \\
\midrule
MHN w/ multiscale, $n$=1,3,2     & 83.2          & 90.7          & 58.1          & \textbf{3.57}               \\
MHN w/ multiscale, $n$=3,1,2     & 82.7          & 90.7          & \textbf{58.2} & 3.61               \\
MHN w/ multiscale, $n$=3,2,1     & 82.6          & 90.8          & 58.0          & 3.60               \\
\midrule
MHN (default)                    & \textbf{83.5} & \textbf{90.8} & 58.1          & 3.58      \\ 
\bottomrule
\end{tabular}%
}
\caption{The impact of the multiscale input on model performance in the TGIF-QA dataset.}
\label{tab:table3}
\end{table}

\subsubsection{Multiscale Information} 
    We first replace the multiscale input with the input at a single temporal scale. Specifically, the multimodal interaction blocks within the RMI module all take the same appearance-motion feature $\mathbf{X}^n$ at a single scale, with $n=1,2,3$ separately. Additionally, for the multiscale input, our default model takes the inputs at scales from $n=1$ to $n=3$. We further change such an input order to consider the inputs at scales of $n=1,3,2$, $n=3,2,1$, and $n=3,1,2$. Results are reported in Table \ref{tab:table3}. We can see that, when the information is provided at a single scale, model performances reduced across all the tasks. Therein, by providing richer frames at higher scales, the Action, Trans, and Count tasks are improved, suggesting their dependencies on richer temporal information. By contrast, while the FrameQA task relies on the visual inference at a single frame, richer frames lead to reduced performances. Furthermore, the change of orders in the multiscale information input does not impact the model performance noticeably, suggesting the incorporation of multiscale information with multilevel model processing does not depend on a rigid matching of orders.

\begin{table}[b]
\centering
\resizebox{0.48\textwidth}{!}{%
\begin{tabular}{lrrrr}
\toprule
Model & Action  & Trans.  & FrameQA  & Count $\downarrow$                             \\ 
\midrule
MHN w/o recurrent connection in RMI   & 82.4          & 89.5          & 57.7          & 3.70  \\
MHN w/o low-level information for PVR  & 82.8          & 90.6          & 57.1          & 3.63  \\
MHN w/o weight sharing in PVR         & \textbf{83.9} & \textbf{90.9} & 57.5 & \textbf{3.57}        
\\
\midrule
MHN (default)                         & 83.5          & 90.8          & \textbf{58.1} & 3.58  \\ 
\bottomrule
\end{tabular}%
}
\caption{The impact of multilevel processing and parallel reasoning on model performance in the TGIF-QA dataset.}
\label{tab:table4}
\end{table}

\subsubsection{Multilevel Processing and Parallel Reasoning} 
    Here, as the first variant, we remove recurrent connections between blocks in the RMI module. That is, each block only receive the appearance-motion feature at a scale as input without the information passed from the previous block, thus the multilevel processing is no longer enabled. Thereon, the PVR module still receives the output from each block. For the second variant, we provide the PVR module only the output of the last block of RMI module. Results are reported in Table \ref{tab:table4}. As we disable the multilevel processing by removing the recurrent connections, model performances decrease across the four tasks. Even with multilevel processing, when we only provide the PVR module with the high-level information, model performances improve a bit but are still lower than our default model. These results show the importance of incorporating multilevel processing with multiscale visual inputs, with our proposed HMN model being a promising implementation for such end. 
    
    From another perspective, we further consider a variant of MHN model that uses separate transformer encoders for the PVR module instead of using a single encoder with weight sharing. While the number of trainable parameters increases from 16.5M (default model) to 22.8M thereon, the model achieves even better performances on Action, Trans., and Count tasks as seen in Table \ref{tab:table4}. In comparison, for methods that have published their codes, PSAC ~\cite{11} (39.1M), HME ~\cite{16} (44.8M), HGA ~\cite{28} (104.1M), L-GCN ~\cite{18} (30.4M), and HCRN ~\cite{19} (42.9M) are all bigger than our model that without weight sharing, but performed worse.

\subsubsection{The Number of Scales} 
    The maximum scale $N$ sets the scope of multiscale sampling, as well as the depth of vision-question interactions within our MHN model. Here, we analyze the impact of setting different values of $N$ on model performance. Since the increase of $N$ by 1 would double the GPU memory consumption, we only experiment with $N\in\{2,3,4\}$. We still use an ascent order of the multiscale information to make the input for our MHN model, \textit{i.e.}, $n=1,2,3$ for $N=3$. Results are reported in Table \ref{tab:table5}. The reactions of performances of different tasks toward the increase of scale $N$ are different. The performances on the FrameQA task improve with larger $N$, showing the robustness of our model for visual inference at a single frame even if more frames are provided. The richer frames also contribute to the Count task. For tasks of Action and Trans., where the informative visual cues exist at specific scales of the video, the scale $N$ is better set to reach a balance between the amount of information provided and model performance. For our MHN model on TGIF-QA dataset, we reach such a balance at $N=3$. We also find that, with $N=2$ where only 3 clips are sampled, our model is able to outperform most state-of-the-art methods reported in Table \ref{tab:table1}.

\begin{table}[t]
\centering
\resizebox{0.48\textwidth}{!}{
\begin{tabular}{lrrrr}
\toprule
Model                  & Action        & Trans.        & FrameQA       & Count. $\downarrow$ \\ 
\midrule
MHN w/ $N=2$           & 82.7          & 90.0          & 58.0          & 3.70               \\
MHN w/ $N=3$ (default) & \textbf{83.5} & \textbf{90.8} & 58.1          & 3.58               \\
MHN w/ $N=4$           & 83.3          & 90.2          & \textbf{58.2} & \textbf{3.55}      \\ 
\bottomrule
\end{tabular}%
}
\caption{The impact of the number of scales on model performance in the TGIF-QA dataset.}
\label{tab:table5}
\end{table}

\section{Conclusion}
    This paper presented a novel Multilevel Hierarchical Network (MHN) with multiscale sampling for accurate VideoQA. In general, MHN enables the incorporation of the multiscale visual information with the multilevel processing capacity of deep learning. Specifically, we designed a Recurrent Multimodal Interaction (RMI) module to use the recurrently-connected multimodal interaction blocks to accommodate the interaction between the visual information and the question across temporal scales. We designed another Parallel Visual Reasoning (PVR) module to adopt a shared transformer encoder layer to process and fuse the multilevel output of RMI for final visual inference. Our extensive experiments conducted on three VideoQA benchmark datasets demonstrated improved performances of our MHN model than previous state-of-the-arts. Our ablation study verified the importance of multiscale visual information for videoQA, and the efficiency and effectiveness of our method on leveraging it.

\section*{Acknowledgments}
    This work is funded by the National Natural Science Foundation of China (62106247). Chongyang Wang is supported by the UCL Overseas Research Scholarship (ORS) and Graduate Research Scholarship (GRS). Yuan Gao is partially supported by the National Key R\&D Program of China (2020YFB1313300).

%% The file named.bst is a bibliography style file for BibTeX 0.99c
\bibliographystyle{named}
\small\bibliography{Paper}

\end{document}